\pdfoutput=1

\documentclass[11pt]{article}

\usepackage{ACL2023}

\usepackage{times}
\usepackage{latexsym}
\usepackage{graphicx}
\usepackage{booktabs}
\usepackage{multirow}
\usepackage{hyperref}
\usepackage{algorithm}
\usepackage{algpseudocode}
\usepackage{enumitem}
\usepackage{subcaption}
\usepackage{listings}
\usepackage{fancyvrb}
\usepackage{xcolor}
\usepackage{amssymb}
\usepackage{pifont}

\newcommand{\cmark}{\ding{51}}%
\newcommand{\xmark}{\ding{55}}%

\newcommand{\pddl}{\texttt{PDDL}\space}

\usepackage[T1]{fontenc}

\usepackage[utf8]{inputenc}

\usepackage{microtype}

\usepackage{inconsolata}

\title{Dynamic Planning with a LLM}

\author{Gautier Dagan \qquad\quad Frank Keller \qquad\quad Alex Lascarides  \\
  School of Informatics \\ 
  University of Edinburgh, UK \\
  \texttt{gautier.dagan@ed.ac.uk}, \texttt{\{keller, alex\}@inf.ed.ac.uk} \\
}

\begin{document}
\maketitle
\begin{abstract}

While Large Language Models (LLMs) can solve many NLP tasks in zero-shot settings, applications involving embodied agents remain problematic.  
In particular, complex plans that require multi-step reasoning become difficult and too costly as the context window grows.
Planning requires understanding the likely effects of one's actions and identifying whether the current environment satisfies the goal state.
While symbolic planners find optimal solutions quickly, they require a complete and accurate representation of the planning problem, severely limiting their use in practical scenarios.
In contrast, modern LLMs cope with noisy observations and high levels of uncertainty when reasoning about a task.  
Our work presents LLM Dynamic Planner (LLM-DP): a neuro-symbolic framework where an LLM works hand-in-hand with a traditional planner to solve an embodied task.
Given action-descriptions, LLM-DP solves Alfworld faster and more efficiently than a naive LLM ReAct baseline.
\end{abstract}

\section{Introduction}

\begin{figure*}[ht]
\centering
\includegraphics[width=\textwidth]{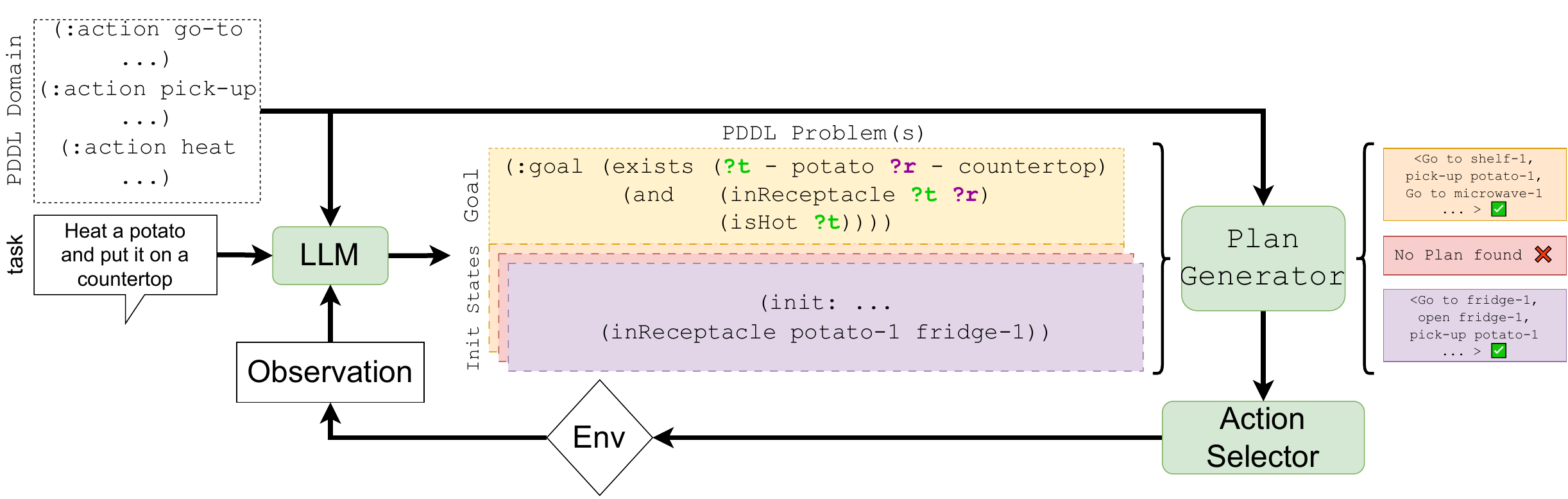}
\caption{LLM Dynamic Planner (LLM-DP). The LLM grounds observations and processes natural language instructions into \pddl to use with a symbolic planner. This model can solve plans for unobserved or previously unknown objects because the LLM generates plausible predicates for relevant objects through semantic and pragmatic inference. Through sampling possible predicates, multiple plans can be found, and an Action Selector decides whether to act, review its understanding of the problem, or ask clarification questions.}
\label{fig:main}
\end{figure*}

Large Language Models (LLMs), like GPT-4 \citep{gpt4}, have proven remarkably effective at various natural language processing tasks, particularly in zero-shot or few-shot settings \cite{fewshot}.
However, employing LLMs in embodied agents, which interact with dynamic environments, presents substantial challenges.
LLMs tend to generate incorrect or spurious information, a phenomenon known as hallucination, and their performance is brittle to the phrasing of prompts \cite{Ji2022SurveyOH}. 
Moreover, LLMs are ill-equipped for naive long-term planning since managing an extensive context over multiple steps is complex and resource-consuming \citep{silver2022,Liu2023LLMPEL}.

Various approaches have aimed to mitigate some of these limitations.
For instance, methods like Chain-of-Thought \citep{Wei2022ChainOT} and Self-Consistency \citep{Wang2022SelfConsistencyIC} augment the context with reasoning traces. 
Other, agent-based approaches, such as ReAct \citep{Yao2022ReActSR}, integrate feedback from the environment iteratively, giving the agent the ability to take `thinking' steps or to augment its context with a reasoning trace.
However, these approaches frequently involve high computational costs due to the iterated invocations of LLMs and still face challenges dealing with the limits of the context window and recovering from hallucinations, which can compromise the quality of the plans.

Conversely, traditional symbolic planners, such as the Fast-Forward planner \citep{ff} or the BFS(f) planner\cite{bfsf}, excel at finding optimal plans efficiently. 
But symbolic planners require problem and domain descriptions as prerequisites \citep{pddl}, which hampers their applicability in real-world scenarios where it may be infeasible to achieve these high informational demands. 
For instance, knowing a complete and accurate description of the goal may not be possible before exploring the environment through actions.

Previous work by \cite{Liu2023LLMPEL} has shown that LLMs can generate valid problem files in the Planning Domain Definition Language (\pddl) for many simple examples.
Yet, the problem of incomplete information remains: agents often need to interact with the world to discover their surroundings before optimal planning can be applied.
Some versions of \pddl have been proposed in the past to deal with probabilities or Task and Motion Planning, such as PPDDL and PDDLStream \cite{younes2004ppddl1, Garrett2018PDDLStreamIS}, but these still assume a human designer encoding the agent's understanding of the domain and the planning problem, rather than the agent learning from interactions.
Therefore, where modern LLMs need minimal information to figure out a task, e.g. through Few-shot or In-Context Learning \citep{Honovich2022InstructionIF, Chen2021MetalearningVL, Min2022RethinkingTR}, traditional planners need maximal information.

In this work, \textbf{we introduce the LLM Dynamic Planner (LLM-DP), a neuro-symbolic framework that integrates an LLM with a symbolic planner to solve embodied tasks}.\footnote{Our code is available at \href{https://github.com/itl-ed/llm-dp}{github.com/itl-ed/llm-dp}}
LLM-DP capitalises on the LLM's ability to understand actions and their impact on their environment and combines it with the planner's efficiency in finding solutions. 
Using domain knowledge, LLM-DP solves the Alfworld test set faster and more efficiently than a LLM-only (ReAct) approach.
The remainder of this paper explores the architecture of LLM-DP, discusses how to combine the strengths of LLMs and symbolic planning and presents potential research avenues for future work in LLM-driven agents.

\section{Related Work}

\textbf{Symbolic Planners} Symbolic planners have been a cornerstone in automated planning and artificial intelligence for decades \cite{strips}. 
Based on formal logic, they operate over symbolic representations of the world to find a sequence of actions that transition from an initial state to a goal state. 
Since the introduction of \pddl \citep{pddl}, the AI planning community has developed an array of efficient planning algorithms.
For example, the Fast-Forward planner (FF) \citep{ff} employs heuristics derived from a relaxed version of the planning problem. 
Similarly, the BFS(f) planner \citep{bfsf} combines breadth-first search and specialised heuristics.
These planners find high-quality or optimal solutions quickly in well-defined domains.
However, their up-front requirement for comprehensive problem and domain descriptions limits their applicability in complex real-world settings where complete information may not be available.

\noindent \textbf{LLMs in Planning and Reasoning} 
In contrast to symbolic planners, LLMs have shown promise in adapting to noisy planning and reasoning tasks through various methods.
Some general approaches such as Chain-of-Thought \citep{Wei2022ChainOT}, Self-Consistency \citep{Wang2022SelfConsistencyIC}, and Reasoning via Planning \citep{Hao2023ReasoningWL} augment the context with a reasoning trace that the LLM generates to improve its final prediction. 
Alternatively, giving access to tools/APIs \cite{Schick2023ToolformerLM, patil2023gorilla}, outside knowledge or databases \cite{Peng2023CheckYF, Hu2023ChatDBAL}, code \cite{Suris2023ViperGPTVI}, and even symbolic reasoners \cite{coupling2023} to enrich an LLM's context and ability to reason. 
The LLM can trigger these external sources of information or logic (through fine-tuning or prompting) to obtain additional context and improve its downstream performance.

\noindent \textbf{Embodied Agents with LLMs} 
In a parallel direction, recent works such as ReAct \citep{Yao2022ReActSR}, Reflexion \citep{shinn2023}, AutoGPT \citep{autoGPT}, and Voyager \citep{Wang2023VoyagerAO}, take an agent-based approach and augment the reasoning process through a closed `while' loop that feeds environment observations back to the LLM.
ReAct \citep{Yao2022ReActSR} allows the LLM agent to either take an action or a `thinking' step.
This allows the LLM to augment its context with its reasoning, which can be seen as agent-driven Chain-of-Thought prompting.
Voyager \citep{Wang2023VoyagerAO} incrementally builds an agent's capabilities from its interactions with the environment and an accessible memory component (skill library).
While many of these works show promising results in building general executable agents in embodied environments \citep{Wang2023VoyagerAO}, they still require many expensive calls to the LLMs, are limited by the LLM's context window, and do not guarantee optimal plans.

\section{Alfworld}

Alfworld \citep{alfworld} is a text-only home environment where an agent is tasked with seven possible tasks, such as interacting with one or more objects and placing them in a specific receptacle. 
At the start of each episode, the goal is given in natural language, and the initial observation does not include the location of any objects.
Therefore an agent must navigate the environment to search for the relevant objects and perform the correct actions.
The possible locations of the environment are known, and the agent can navigate to any receptacle by using a `go to' action.
However, since none of the objects' locations are initially observed, the agent must be able to plan around uncertainty, estimate where objects are likely to be observed and adjust accordingly.

\section{LLM-DP}

To tackle an embodied environment like Alfworld, we introduce the Large Language Model Dynamic Planner (LLM-DP), which operates as a closed-loop agent.
LLM-DP uses a combination of language understanding and symbolic reasoning to plan and solve tasks in the simulated environment.
The model tracks a World State $\mathcal{W}$ and beliefs $\mathcal{B}$ about predicates in the environment, uses an LLM to translate the task description into an executable goal state and samples its beliefs to generate plausible world states. 
We describe the working of the LLM-DP agent as pseudo-code in Algorithm~\ref{alg:llmdp}.

\begin{algorithm}[t]
    \caption{LLM-DP Pseudo-code}\label{alg:llmdp}
    \begin{algorithmic} 
        \Require LLM, PG, AS, Domain, $task$, $obs_0$
        \State $goal \gets$ LLM$($Domain$, task)$
        \State $\mathcal{W}, \mathcal{B} \gets $observe$(goal, obs_0)$
        \While{$goal$ not reached}
            \State $plans \gets \emptyset$
            \For{$i$ in $N$}
                \State $w_{belief}\gets $LLM$(\mathcal{B}, \mathcal{W})$
                \State $plans \gets $PG$(w_{belief} \bigcup \mathcal{W})$
            \EndFor
            \State $action \gets $AS$(plans)$
            \State $obs \gets $Env$(action)$
            \State $\mathcal{W}, \mathcal{B} \gets $observe$(action, obs)$
        \EndWhile
    \end{algorithmic}
\end{algorithm}

\begin{table*}
\centering
\begin{subtable}{\textwidth}
\centering
\begin{tabular}{lrrrrrr|r|r}
\toprule
\multicolumn{8}{c}{Average Accuracy (\%)} & \\
Model & clean & cool & examine & heat & put & puttwo & overall ($\uparrow$) & LLM Tokens ($\downarrow$) \\
\midrule
\texttt{LLM-DP} & 0.94 & 1.00 & 1.00 & 0.87 & 1.00 & 0.94 & 0.96 & 633k \\
\texttt{LLM-DP-random} & 0.94 & 1.00 & 1.00 & 0.87 & 0.96 & 1.00 & 0.96 & 67k  \\
\texttt{ReAct} \citep{Yao2022ReActSR} & 0.61 & 0.81 & 0.89 & 0.30 & 0.79 & 0.47 & 0.64 & ---*\\
\texttt{ReAct} (ours) & 0.35 & 0.90 & 0.33 & 0.65 & 0.71 & 0.29 & 0.54 & 9.16M  \\
\midrule
\end{tabular}
\caption{The average accuracy and number of LLM Tokens processed (context + generation) for each model. *Not reported.}
\end{subtable}

\begin{subtable}{\textwidth}
\centering
\begin{tabular}{lrrrrrr|r}
\multicolumn{8}{c}{Average Episode Length} \\
Model & clean & cool & examine & heat & put & puttwo & overall ($\downarrow$) \\
\midrule
\texttt{LLM-DP} & 12.00 & 13.67 & 12.06 & 12.30 & 12.75 & 17.59 & 13.16 \\
\texttt{LLM-DP-random} & 15.06 & 17.14 & 10.56 & 14.04 & 14.62 & 18.94 & 15.02 \\
\texttt{ReAct} (ours) & 25.10 & 9.86 & 21.67 & 14.70 & 15.33 & 24.94 & 18.69 \\
\bottomrule
\end{tabular}
\caption{The average episode length for each model, where the length of an episode denotes how many actions the agent has taken or attempted to take to complete a task. We do not count the `thinking' action of \texttt{ReAct} as an action in this metric.}
\end{subtable}
\caption{Summary of model performance on the Alfword test set. \texttt{LLM-DP} and \texttt{LLM-DP-random} differ in the sampling strategy of the belief. \texttt{LLM-DP} uses an LLM to generate $n=3$ plausible world states, while \texttt{LLM-DP-random} randomly samples $n=3$ plausible world states.}
\label{table:main}
\end{table*}

\subsection{Assumptions}
\label{sec:assumptions}

We make several simplifying assumptions when applying the LLM-DP framework to Alfworld:

\begin{enumerate}[itemsep=0mm]
    \item \textbf{Known action-descriptions and predicates}: Our input to the planner and the LLM requires the \pddl domain file, which describes what actions can be taken, their pre- and post-conditions, and what predicates exist.
    \item \textbf{Perfect observations}: The Alfworld environment provides a perfect textual description of the current location. 
    This observation also contains the intrinsic attributes of observed objects and receptacles, such as whether or not a given receptacle can be opened.
    \item \textbf{Causal Environment}: changes in the environment are entirely caused by the agent.
    \item \textbf{\textit{Valid} actions always succeed}
\end{enumerate}

\subsection{Generating the Goal State}
\label{sec:goal}

LLM-DP uses an LLM to generate a \pddl goal, given the natural language instruction ($task$) and the valid predicates defined by the \pddl domain file.
Figure~\ref{fig:main} shows an example task converted to a valid \pddl goal.
For each episode, we use a set of three in-context examples that are fixed for the entire evaluation duration.
We use the OpenAI \texttt{gpt-3.5-turbo-0613} LLM model with a temperature of 0 in all our LLM-DP experiments.

\subsection{Sampling Beliefs}

We parse the initial scene description into a structured representation of the environment $\mathcal{W}$ and a set of beliefs $\mathcal{B}$.
The internal representation of the world $\mathcal{W}$ contains all \textit{known} information, for instance, all receptacles (possible locations) in the scene from the initial observation and their intrinsic attributes are known (i.e. a fridge holds the \texttt{isFridge} predicate).
Whereas the set of beliefs $\mathcal{B}$ are a set of possible valid predicates that can be true or false and which the model does not have enough information to disambiguate.
In Alfworld, the objects' locations are unknown; therefore, the set of possible predicates for each object includes all possible locations.

LLM-DP uses stored observations $\mathcal{W}$, beliefs $\mathcal{B}$ and an LLM to construct different planning problem files in \pddl.
A \pddl problem file includes the objects observed (\texttt{:objects}), a representation of the current state (\texttt{:init}) of the world and the object attributes, and the goal to be achieved (\texttt{:goal}).
The goal is derived from the LLM (Section~\ref{sec:goal}), while the objects and their attributes are obtained from $\mathcal{W}$ (observations) and the beliefs the $\mathcal{B}$ has about the objects.

Since $\mathcal{B}$ includes possible predicates which are unknown, we sample from $\mathcal{B}$ using an LLM to obtain $w_{belief}$.
For instance, our belief could be that \texttt{(inReceptacle tomato ?x)} where \texttt{?x} can be \texttt{countertop}, \texttt{cabinet}, \texttt{fridge}, etc.
Since we want to condition the sampling of where the \texttt{tomato} can appear, we pass the known world state $\mathcal{W}$ along with the predicate (in this case \texttt{inReceptacle}) and its options to the LLM.
This sampling leverages the LLM to complete a world state and is extendable to any unknown predicate from which a set of beliefs can be deduced.
We also compare LLM sampling with random sampling (\texttt{llmdp-random}).

We describe our likely world state as the union between a sampled set of beliefs and the known world state $w_{belief} \bigcup \mathcal{W}$.
Then sampling $i=1,.., N$ different sets of beliefs during the planning loop, we obtain $N$ likely world states.
Finally, we convert each likely world state to lists of predicates to interface with the \pddl planner.

\subsection{Plan Generator}

Upon constructing the different \pddl problems, the agent uses a Plan Generator (PG) to solve each problem and obtain a plan.
We use the BFS(f) solver \citep{bfsf} implemented as an executable by LAPKT \citep{lapkt}.
A generated plan is a sequence of actions, where each action is represented in a symbolic form, which, if executed, would lead to the goal state from the initial state.

\subsection{Action Selector}

The Action Selector (AS) module decides the agent's immediate next action. 
It takes the planner's output, a set of plans, and selects an action from them.
In our Alfworld experiments, the Action Selector simply selects the shortest plan returned.
If no valid plans are returned, all sampled states were satisfying goal states, there is a mistake with the constructed domain/problem files, or the planner has failed to find a path to the goal.
In the first case, we re-sample random world states and re-run the planners once.

We also propose exploring different strategies when valid plans cannot be found.
For instance, similarly to self-reflection \cite{shinn2023}, the Action Selector could prompt an update in the agent's belief about the world state if none of generated problem descriptions are solvable.
The Action Selector could also interact with a human teacher or oracle to adjust its understanding of the environment (problem) or its logic (domain).

\subsection{Observation Processing}

LLM-DP uses the result of each action to update its internal state representation. 
It uses the symbolic effects of the action to infer changes in the state of the objects and receptacles.
Then it integrates the information from the new observation, which might reveal additional details not directly inferred from the action itself.
For instance, opening an unseen drawer might reveal new objects inside.
Observing also updates the beliefs -- if an object is observed at a location, it cannot be elsewhere, but if an object is not observed at a location, it cannot be there.
Observations incorporate beliefs into $\mathcal{W}$.

If the agent detects new information from the scene - such as discovering new objects - it triggers a re-planning process.
The agent then generates a new set of possible \pddl problems using the updated state representation and corresponding plans using the Plan Generator.
This approach is similar to some Task and Motion Planning (TAMP) methods \citep{Garrett2018PDDLStreamIS, Chen2023AutoTAMPAT}, enabling the agent to adapt to environmental changes and unexpected outcomes of actions.

\section{Results}

We contrast the LLM-DP approach with ReAct (LLM-only baseline) from the original implementation by \citet{Yao2022ReActSR}.
Since we use a different backbone LLM model (\texttt{gpt-3.5-turbo} rather than \texttt{text-davinci-002}) than the ReAct baseline for cost purposes, we also reproduce their results using \texttt{gpt-3.5-turbo} and adapt the ReAct prompts to a chat format.

As shown in Table~\ref{table:main}, LLM-DP solves Alfworld almost perfectly ($96\%$) compared to our baseline reproduction of ReAct ($53\%$).
The LLM-DP can translate the task description into an executable \pddl goal $97\%$ of the time, but sampling reduces the accuracy further when it fails to select a valid set of possible world states -- for instance, by sampling states where the goal is already satisfied.

We note, that the ReAct baseline makes different assumptions about the problem; while it does not require a domain file containing the action-descriptions and object predicates, it uses two separate human-annotated episodes per example to bootstrap its in-context logic.
ReAct also switches out which examples to use in-context based on the type of task, such that two examples of the same type of task being solved are always shown.
We also find that our reproduction of ReAct is worse than the original and attribute this to the \texttt{gpt-3.5-turbo} model being more conversational than \texttt{text-davinci-002}, and thus less likely to output valid actions as it favours fluency over following the templated action language.

We also measure the length of each successful episode and find that LLM-DP reaches the goal state faster on average (13.16 actions) versus ReAct (18.69 actions) and a random search strategy (15.02 actions).
The Average Episode Length measures the number of actions taken in the environment and how efficient the agent is.

\section{Conclusion}

The LLM-DP agent effectively integrates language understanding, symbolic planning, and state tracking in a dynamic environment. 
It uses the language model to understand tasks and scenes expressed in natural language, constructs and solves planning problems to decide on a course of action, and keeps track of the world state to adapt to changes and make informed decisions. 
This workflow enables the agent to perform complex tasks in the Alfworld environment, making it a promising approach for embodied tasks that involve language understanding, reasoning, and decision-making.

LLM-DP offers a cost and efficiency trade-off between a wholly symbolic solution and an LLM-only model.
The LLM's semantic knowledge of the world is leveraged to translate the problem into \pddl while guiding the search process through belief instantiation.
We find that not only is LLM-DP cheaper, on a per-token comparison, but it is also faster and more successful at long-term planning in an embodied environment.
LLM-DP validates the need for LLM research to incorporate specialised tools, such as \pddl solvers, in embodied agents to promote valid

Despite these promising results, numerous topics and unresolved issues remain open for future investigation. 
Key among these is devising strategies to encode the world model and belief, currently handled symbolically, and managing uncertain observations --- particularly from an image model --- along with propagating any uncertainty to the planner and Action Selector. 
We intentionally kept the Action Selector simple for our experiments, but future work may also explore different strategies to encourage self-reflection within the agent loop. 
For instance, if all plans prove invalid, beliefs may be updated, or it might indicate an incorrect domain definition.
Such instances may necessitate agents to interact with an instructor who can provide insights about action pre-conditions and effects.
This direction could lead us from a static domain file towards an agent truly adaptable to new environments, fostering continual learning and adaptation.

\section*{Acknowledgements}

This work was supported in part by the UKRI Centre for Doctoral Training in Natural Language Processing, funded by the UKRI (grant EP/S022481/1) at the University of Edinburgh, School of Informatics and School of Philosophy, Psychology \& Language Sciences and by the UKRI-funded TAS Governance Node (grant number EP/V026607/1).

\bibliography{anthology,custom}
\bibliographystyle{acl_natbib}

\appendix

\begin{table}[ht]
\begin{tabular}{lll}
 & SR & EL \\
\toprule 
LLM-DP (n=3) & 0.96 & 13.16 \\
LLM-DP (n=3) - fallback & 0.92 & 12.80 \\
LLM-DP (n=5) & 0.96 & 12.54 \\
LLM-DP (n=5) - fallback & 0.94 & 12.24
\end{tabular}
\caption{We compare the average Success Rate (SR) and average Episode Length (EL) for different sampling sizes $n$ and with or without a fallback to random sampling. The random sampling fallback affects the success rate as the LLM sampler can more often sample $n$ world states which are already satisfied.
However as $n$ increases, we see that it becomes more likely for the sampling procedure to at find at least one plan, and therefore the SR increases when no fallback (- fallback) is used.
}
\end{table}
\label{table:ablation}
\section{Prompts and Few-shot details}
\label{sec:appendix_sampling}

See Table~\ref{table:sys-llmdp} and Table~\ref{table:few-llmdp} for LLM-DP prompts used.

\begin{table*}
\begin{small}
\begin{verbatim}
(define (domain alfred)
(:predicates
    (isReceptacle ?o - object) ; true if the object is a receptacle
    (atReceptacleLocation ?r - object) ; true if the robot is at the receptacle location
    (inReceptacle ?o - object ?r - object) ; true if object ?o is in receptacle ?r
    (openable ?r - object) ; true if a receptacle is openable
    (opened ?r - object) ; true if a receptacle is opened
    (isLight ?o - object) ; true if an object is light source
    (examined ?o - object ?l - object) ; whether the object has been looked at with light
    (holds ?o - object) ; object ?o is held by robot
    (isClean ?o - object) ; true if the object has been cleaned in sink
    (isHot ?o - object) ; true if the object has been heated up
    (isCool ?o - object) ; true if the object has been cooled
    (isSink ?o - object) ; true if the object is a sink
    (isMicrowave ?o - object) ; true if the object is a microwave
    (isFridge ?o - object) ; true if the object is a fridge
))
\end{verbatim}
\end{small}
\caption{System Prompt used by \texttt{gpt-3.5-turbo} for generating the \texttt{:goal} in LLM-DP}
\label{table:sys-llmdp}
\end{table*}

\begin{table*}
\begin{small}
\begin{verbatim}
Your task is to: put a clean plate in microwave.   
(:goal
(exists (?t - plate ?r - microwave)
(and (inReceptacle ?t ?r)
(isClean ?t)
)))

Your task is to: examine an alarmclock with the desklamp",
(:goal
(exists (?t - alarmclock ?l - desklamp)
(and (examined ?t ?l) (holds ?t)
)))

Your task is to: put two cellphone in bed
(:goal
(exists (?t1 - cellphone ?t2 - cellphone ?r - bed)
(and (inReceptacle ?t1 ?r)
(inReceptacle ?t2 ?r)
(not (= ?t1 ?t2))
)))
\end{verbatim}
\end{small}
\caption{Fixed Few-shot examples used by \texttt{gpt-3.5-turbo} for  generating the \texttt{:goal} in LLM-DP}
\label{table:few-llmdp}
\end{table*}

\section{ReAct}
\label{sec:appendix_react}
\subsection{Reproduction with Chat Model}

We slightly modify the `system' prompt of the original ReAct (see Table~\ref{table:sys-react}) to guide the model away from its conversational tendencies. 
\texttt{gpt-3.5-turbo} apologises significantly more than the  \texttt{text-davinci-002} model, and we found that it would often get stuck in loops of apologising.
We also modify the code so that we replace all generated instances of `in' and `on' with `in/on' if the model did not generate it correctly, since Alfworld expects `in/on' but \texttt{gpt-3.5-turbo} tends to generate only the correct preposition.
Without these changes, ReAct would be significantly worse than our reported metric.

\begin{table*}
\begin{small}
\begin{verbatim}
Interact with a household to solve a task.
Only reply with > followed by the action to take or 'think'.
Do not apologize.
Follow the format of the two examples below.
\end{verbatim}
\end{small}
\caption{System Prompt used by \texttt{gpt-3.5-turbo} in our reproduction of ReAct}
\label{table:sys-react}
\end{table*}

\begin{table*}[h]
\centering
\begin{tabular}{|l|l|}
\toprule
\begin{minipage}{0.5\textwidth}
\begin{small}
\begin{Verbatim}[commandchars=\\\{\}]
task: put some peppershaker on drawer.
Generated: 
(:goal
    (exists (?t - peppershaker ?r - drawer)
        (inReceptacle ?t ?r)
))

VALID \cmark
\end{Verbatim}
\end{small}
\end{minipage} & 
\begin{minipage}{0.5\textwidth}
\begin{small}
\begin{Verbatim}[commandchars=\\\{\}]
task: put a clean mug in coffeemachine.
Generated: 
(:goal
    (exists (?t - mug ?r - coffeemachine)
        (and (inReceptacle ?t ?r)
             (isClean ?t)
)))

VALID \cmark
\end{Verbatim}
\end{small}
\end{minipage} \\
\midrule
\begin{minipage}{0.5\textwidth}
\begin{small}
\begin{Verbatim}[commandchars=\\\{\}]
task: put two cd in safe.
Generated: 
(:goal
    (exists (?t1 - cd ?t2 - cd ?r - safe)
        (and (inReceptacle ?t1 ?r)
             (inReceptacle ?t2 ?r)
             (not (= ?t1 ?t2))
)))
VALID \cmark
\end{Verbatim}
\end{small}
\end{minipage} & 
\begin{minipage}{0.5\textwidth}
\begin{small}
\begin{Verbatim}[commandchars=\\\{\}]
task: heat some mug and put it in coffeemachine.
Generated: 
(:goal
    (exists (?m - mug ?c - coffeemachine)
        (and \textcolor{red}{(isReceptacle ?m)}
             (isHot ?m)
             (inReceptacle ?m ?c)
)))
INVALID \xmark
\end{Verbatim}
\end{small}
\end{minipage} \\
\bottomrule
\end{tabular}
\caption{Sample of generated \pddl goals from LLM-DP. The generation gets confused by the semantics of `receptacle' and identifies a mug as a receptacle. While it is true that a mug is a receptacle, in our defined logic, receptacles are fixed, immovable objects which can contain other objects and therefore, a mug is not a Receptacle which leads the planning to fail subsequently.}
\label{table:generated-goals}
\end{table*}

\section{LLM-DP}
\label{sec:appendix_llmdp}

\subsection{Generated Goal Examples}
See Table~\ref{table:generated-goals} for examples of generated goals, both valid and invalid.

\subsection{Varying $n$}

See Table~\ref{table:ablation} for results when different varying $n$ and fallback. Fallback is when no plans are sampled successfully through the LLM, LLM-DP re-samples $n$ plans randomly.

\end{document}